\batchmode
\documentclass{llncs}

\usepackage{fancyvrb}
\usepackage[dvipdfm]{graphicx}

\title{Unsupervised Grammar Induction in a Framework of Information
Compression by Multiple Alignment, Unification and Search} 

\author{J Gerard Wolff}

\institute{CognitionResearch.org.uk,\\
Telephone: +44(0)1248 712962,\\
Email: jgw@cognitionresearch.org.uk.}

\begin{document}

\maketitle

\begin{abstract}

This paper describes a novel approach to grammar induction that has been developed within a framework designed to integrate learning with other aspects of computing, AI, mathematics and logic. This framework, called {\em information compression by multiple alignment, unification and search} (ICMAUS), is founded on principles of Minimum Length Encoding pioneered by Solomonoff and others. Most of the paper describes SP70, a computer model of the ICMAUS framework that incorporates processes for unsupervised learning of grammars. An example is presented to show how the model can infer a plausible grammar from appropriate input. Limitations of the current model and how they may be overcome are briefly discussed.

\end{abstract}

\section{Introduction}

This paper describes a novel approach to unsupervised grammar induction that has been developed within a research programme whose overarching goal is the {\em integration} of diverse functions---learning, recognition, reasoning and others---within one relatively simple framework. This has had a substantial impact on the way in which the learning processes are organised.

The new framework called {\em information compression by multiple alignment, unification and search} (ICMAUS) originated in earlier research developing the SNPR model of grammar induction \cite{wolff_1988,wolff_1982}. Without supervision, the SNPR model successfully learns artificial context-free phrase-structure grammars (CF-PSGs) using a technique of `hierarchical chunking' combined with a search for disjunctive (part of speech) categories and processes for generalising grammatical rules and correcting over-generalisations.

In the ICMAUS programme, the aim has been to match or exceed these capabilities within a system that has been generalised to model a range of other aspects of computing, AI, mathematics and logic. It became apparent at an early stage that this would mean a radical reorganisation of the SNPR model. In the ICMAUS framework a concept of {\em multiple alignment}---to be described---has replaced hierarchical chunking as the predominant mode of organisation. With this new orientation, the system provides an interpretation for concepts in computing, mathematics and logic and it has a range of AI capabilities described in \cite{wolff_icmaus_overview} and earlier papers cited there. The present paper describes how the system has been developed for unsupervised learning of grammars.

A much fuller account of the research described here may be found in \cite{wolff_unsupervised_learning}, available from http://www.cognitionresearch.org.uk/papers/ul/ul.htm.

\subsection{Relationship with Other Research on Grammar Induction}

This research extends the tradition of distributional linguistics pioneered by \cite{harris_1951,fries_1952} and others.

At the heart of ICMAUS system are principles of Minimum Length Encoding (MLE) pioneered by \cite{solomonoff_1964} (see also \cite{li_vitanyi_1997}). In this framework, grammar induction is conceived as a process of optimisation rather than a process of identifying a target grammar `in the limit' as postulated by \cite{gold_1967}. In the MLE framework, there is no target grammar, merely a process of searching for grammars that are `good' in terms of MLE principles.

Recent studies that are, perhaps, most closely related to the present research include: \cite{adriaans_et_al_2000,allison_wallace_yee_1992,clark_2001,denis_2001,henrichsen_2002,johnson_reizler_2002,klein_manning_2001,nevill-manning_witten_1997,oliveira_sv_1996,rapp_et_al_1994,solan_etal_2002,van_zaanen_2002,van_zaanen_thesis_2002,watkinson_manandhar_2001}. Not all of these studies have adopted MLE principles but they deal with issues and processes that relate to the present research. The idea of combining learning with parsing---to be described---has also been developed by Nakamura (see \cite{nakamura_ishiwata_2000} and this workshop).

Compared with other work on unsupervised learning of grammar-like structures, the most distinctive features of the ICMAUS research are:

\begin{itemize}

\item The integration of learning with other areas of AI, computation, mathematics and logic.

\item \sloppy The multiple alignment concept as it has been developed in the ICMAUS framework, described below. There is, however, a clear affinity with `alignment-based learning' \cite{van_zaanen_2002,van_zaanen_thesis_2002}.

\end{itemize}

\section{The ICMAUS Framework}\label{ICMAUS_section}

In the ICMAUS framework, {\em all} knowledge is stored as {\em patterns}: arrays of symbols in one or two dimensions.\footnote{In work to date, the focus has been on one-dimensional patterns.} Despite the simplicity of this format, it is possible within the ICMAUS system to represent several different kinds of knowledge including context-free and context sensitive grammars, networks, trees, if-then rules and others.

Given the generality of this format for knowledge, the learning techniques described in this paper are relevant to the learning of {\em any} kind of knowledge, not just `grammars', narrowly conceived.

The ICMAUS framework is intended as an abstract model of {\em any} kind of system for computing or cognition, either natural or artificial. In broad terms, the system works by receiving `New' information from its environment and transferring it to a repository of `Old' information. At the same time, it tries to compress the information as much as possible by finding patterns that match each other and merging or `unifying' patterns that are the same. In these broad terms it is similar to a ZIP program but it differs in the thoroughness of the search for `good' unifications of patterns and in the `multiple alignment' concept, to be described.

\subsection{Multiple Alignment}\label{multiple_alignment_section}

The concept of {\em multiple alignment} in the ICMAUS framework has been borrowed from the field of bio-informatics and adapted as described in \cite{wolff_icmaus_overview}.

An example of an ICMAUS multiple alignment is shown in Figure \ref{alignment_1}. Row 0 contains the New pattern `o n e o f t h e m d o e s' and all the other rows contain Old patterns, one pattern per row. By convention, the New pattern is always shown in row 0 but otherwise the assignment of patterns to rows is entirely arbitrary.

\begin{figure}[!hbt]
\fontsize{06.00pt}{07.20pt}
\begin{center}
\begin{BVerbatim}
0                             o n e               o f            t h e m                d o e s   0
                              | | |               | |            | | | |                | | | |  
1                             | | |               | |   < N Np 0 t h e m >              | | | |   1
                              | | |               | |   | |              |              | | | |  
2                             | | |   < Q 0 < P   | | > < N              > >            | | | |   2
                              | | |   | |   | |   | | |                    |            | | | |  
3                             | | |   | |   < P 2 o f >                    |            | | | |   3
                              | | |   | |                                  |            | | | |  
4                             | | |   | |                                  |   < V Vs 1 d o e s > 4
                              | | |   | |                                  |   | | |            |
5 S Num     ; < NP            | | |   | |                                  | > < V |            > 5
     |      | | |             | | |   | |                                  | |     |             
6    |      | | |    < N Ns 3 o n e > | |                                  | |     |              6
     |      | | |    | | |          | | |                                  | |     |             
7    |      | < NP 0 < N |          > < Q                                  > >     |              7
     |      |            |              |                                          |             
8   Num SNG ;            Ns             Q                                          Vs             8
\end{BVerbatim}
\end{center}
\caption{A multiple alignment with `o n e o f t h e m d o e s' in New and patterns representing grammatical rules in Old.}
\label{alignment_1}
\end{figure}

Apart from the pattern in row 8, the patterns from Old in this example are like re-write rules in a CF-PSG with the re-write arrow omitted. If we ignore row 8, the alignment shown in Figure \ref{alignment_1} is very much like a conventional parsing, marking the main components of the sentence: words and phrases and the sentence pattern itself (shown in row 5).

Row 8 shows how the `discontinuous' dependency that exists between the singular noun in the subject of the sentence (`Ns') and the singular verb (`Vs') can be marked within the alignment in a relatively direct manner. Despite the simplicity of the format for representing knowledge, the formation of multiple alignments enables the system to express `context sensitive' aspects of language and other kinds of knowledge.

In each Old pattern there are two kinds of symbols: {\em ID-symbols} like `$<$', `N', `Np', `0' and `$>$' in `$<$ N Np 0 t h e m $>$' serve to identify the pattern and the remaining symbols (`t h e m' in this example) are {\em C-symbols} that represent the contents or substance of the pattern.

Much more detail, with many more examples, may be found in \cite{wolff_2000}.

\section{SP70}\label{SP70_section}

All the main components of the ICMAUS framework outlined in Section \ref{ICMAUS_section} are now realised within the SP70 software model (version 9.2). The model is able to abstract plausible grammars from sets of simple sentences without prior knowledge of word segments or the classes to which they belong, and the computational complexity of the model appears to be acceptable (Section \ref{computational_complexity}). However, in its current form, the model has at least two significant shortcomings and some other deficiencies, discussed briefly in Section \ref{discussion_section}. 

\subsection{Objectives}

In the development of this model, the main problems that have been addressed are:

\begin{itemize}

\item How to identify significant segments in the `corpus' of raw data when the boundary between one segment and the next is not marked explicitly.

\item How to identify disjunctive classes of syntactically-equivalent segments (e.g., `nouns', `verbs' and `adjectives').

\item How to combine the learning of segmental structure with the learning of disjunctive classes.

\item How to learn segments and disjunctive classes through two or more levels of abstraction.

\item \sloppy How to generalize grammatical rules beyond the data and how to correct over-generalizations without feedback from a `teacher' or the provision of `negative' samples or the grading of the data from `easy' to `hard' ({\em cf.} \cite{gold_1967}).

\end{itemize}

Solutions to these problems were found in the SNPR model \cite{wolff_1988,wolff_1982} but, as noted earlier, the organisation of this model is quite unsuited to the wider goals of the present research---integration of diverse functions within one framework. The SP70 model (v.~9.2) provides solutions to the first three problems and partial solutions to the fourth and fifth problems. Further development is planned as indicated in Section \ref{discussion_section}, below.

\subsection{Overall Structure of the Model}

Figure \ref{SP70_figure} shows the high-level organisation of the SP70 model.

\begin{figure}[!hbt]
\begin{center}
\fontsize{08.00pt}{09.60pt}
\begin{BVerbatim}
SP70()
{
     1 Read a set of patterns into New. Old is initially empty.
     2 Compile an alphabet of symbol types in New and, for each type,
          find its frequency of occurrence and the number of bits
          required to encode it (using the Shannon-Fano-Elias method).
     3 While (there are unprocessed patterns in New)
     {
          3.1 Identify the first or next pattern from New as the
               `current pattern from New' (CPFN).
          3.2 Apply the function CREATE_MULTIPLE_ALIGNMENTS() to
               create multiple alignments, each one between the
               CPFN and one or more patterns from Old.
          3.3 During 3.2, the CPFN is copied into Old, one symbol
               at a time, in such a way that the CPFN can be
               aligned with its copy but that any one symbol in
               the CPFN cannot be aligned with the corresponding
               symbol in the copy.
          3.4 Sort the alignments formed by this function in order
               of their compression scores and select the best
               few for further processing.
          3.5 Process the selected alignments with the function
               DERIVE_PATTERNS(). This function derives encoded
               patterns from alignments and adds them to Old.
     }

     4 Apply the function SIFTING_AND_SORTING() to create one or
          more alternative grammars for the patterns in New, each
          one scored in terms of MLE principles. Each grammar is
          a subset of the patterns in Old.
}
\end{BVerbatim}
\end{center}
\caption{The organisation of SP70. The workings of the functions {\em create\_multiple\_alignments()}, {\em derive\_patterns()} and {\em sifting\_and\_sorting()} are explained in the text.}
\label{SP70_figure}
\end{figure}

The function {\em create\_multiple\_alignments()} referred to in Figure \ref{SP70_figure} creates zero or more multiple alignments, each one comprising the current pattern from New (CPFN) and one or more patterns from Old. This function is essentially the same as the main component of the SP61 model, described quite fully in \cite{wolff_2000}. Readers are referred to this source for a more detailed description of how multiple alignments are formed in the ICMAUS framework.

\subsection{Deriving Patterns from Alignments}\label{derive_patterns_section}

In operation 3.5 in Figure \ref{SP70_figure}, the {\em derive\_patterns()} function is applied to a selection of the best alignments formed and, in each case, it looks for sequences of unmatched symbols within the alignment and also sequences of matched symbols.

Consider the alignment shown in Figure \ref{alignment_2}. From an alignment like that, the function finds the unmatched sequences `g i r l' and `b o y' and, within row 1, it also finds the matched sequences `t h a t' and `r u n s'. With respect to row 1, the focus of interest is the matched and unmatched sequences of C-symbols---ID-symbols are ignored.

\begin{figure}[!hbt]
\begin{center}
\begin{BVerbatim}
0        t h a t g i r l r u n s   0
         | | | |         | | | |  
1 < %1 9 t h a t b o y   r u n s > 1
\end{BVerbatim}
\end{center}
\caption{A simple alignment from which other patterns may be derived.}
\label{alignment_2}
\end{figure}

A copy of each of the four sequences is made, ID-symbols are added to each copy and the copy is added to Old. In addition, another `abstract' pattern is made that records the sequence of matched and unmatched patterns within the alignment. The result in this case is five patterns like those shown in Figure \ref{patterns_figure_1}.

\begin{figure}[!hbt]
\begin{center}
\begin{BVerbatim}
< %7 12 t h a t >
< %9 14 b o y >
< %9 15 g i r l >
< %8 13 r u n s >
< %10 16 < %7 > < %9 > < %8 > >
\end{BVerbatim}
\end{center}
\caption{Patterns derived from the alignment shown in Figure \ref{alignment_2}.}
\label{patterns_figure_1}
\end{figure}

It should be clear that the set of patterns in Figure \ref{patterns_figure_1} is, in effect, a simple grammar for the two sentences in Figure \ref{alignment_2}, with patterns representing grammatical rules in much the same style as those shown in Figure \ref{alignment_1}. The abstract pattern `$<$ \%10 220 $<$ \%7 $>$ $<$ \%9 $>$ $<$ \%8 $>$ $>$' describes the overall structure of this kind of sentence with slots that may receive individual words at appropriate points in the pattern.

Notice how the symbol `\%9' serves to mark `b o y' and `g i r l' as alternatives in the middle of the sentence. This is a grammatical class in the tradition of distributional or structural linguistics (see, for example, \cite{harris_1951,fries_1952}).

\subsection{Sifting and Sorting of Patterns}\label{sifting_and_sorting_section}

In the example just shown, all the patterns derived from the alignment are `correct'. But in many cases, patterns that are derived in this way and added to Old are `wrong'. The wrong patterns are weeded out in the {\em sifting\_and\_sorting()} stage of processing (operation 4 in Figure \ref{SP70_figure}), where the system develops one or more alternative grammars for the patterns in New in accordance with MLE principles. Figure \ref{sifting_and_sorting_figure} shows the overall structure of the {\em sifting\_and\_sorting()} function.

\begin{figure}[!hbt]
\begin{center}
\fontsize{08.00pt}{09.60pt}
\begin{BVerbatim}
SIFTING_AND_SORTING()
{
     1 For each pattern in Old, set its frequency of occurrence to 0.
     2 While (there are still unprocessed patterns in New)
     {
          2.1 Identify the first or next pattern from New as the CPFN.
          2.2 Apply the function CREATE_MULTIPLE_ALIGNMENTS() to
               create multiple alignments, each one between the CPFN
               and one or more patterns from Old.
          2.3 From amongst the best of the multiple alignments formed,
               select `full' alignments in which all the symbols of
               the CPFN are matched and all the C-symbols are
               matched in each pattern from Old.
          2.4 For each pattern from Old, count the maximum number of
               times it appears in any one of the full alignments
               selected in operation 2.3. Add this count to the
               frequency of occurrence of the given pattern.
     }
     3 Compute frequencies of symbol types and their encoding costs.
          From these values, compute encoding costs of patterns in
          Old and new compression scores for each of the full
          alignments created in operation 2.
     4 Using the alignments created in 2 and the values computed in
          operation 3, COMPILE_ALTERNATIVE_GRAMMARS().
}
\end{BVerbatim}
\end{center}
\caption{The organisation of the {\em sifting\_and\_sorting()} function. The {\em compile\_alternative\_grammars()} function is described in the text.}
\label{sifting_and_sorting_figure}
\end{figure}

\subsubsection{Compiling a Set of Alternative Grammars}\label{compile_grammars}

\sloppy A set of alternative grammars for the patterns in New that are good in terms of MLE principles are derived (in the {\em compile\_alternative\_grammars()} function) in operation 4 of Figure \ref{sifting_and_sorting_figure}. Each grammar is a subset of the patterns that have been added to Old during operation 3 of Figure \ref{SP70_figure}.

The process of compiling good grammars is essentially a hill-climbing search through the abstract space of alternative grammars, trying to minimise $(G + E)$ for each grammar, where $G$ is the size of the given grammar (in bits) and $E$ is the size of all the New patterns (in bits) after they have been encoded in terms of the grammar. Minimising $(G + E)$ is, of course, the central idea in grammar induction using MLE principles. In what follows, $(G + E)$ is abbreviated as $T$.

The grammars are built in stages, at first trying to minimise $T$ for the first New pattern alone, then trying to minimise $T$ for the first and second New pattern, followed by the first, second and third, and so on.

\section{Computational Complexity}\label{computational_complexity}

In a serial processing environment, the time complexity of SP70 is approximately O$(N^2)$ where $N$ is the number of patterns in New. In a parallel processing environment, the time complexity may approach O$(N)$, depending on how well the parallel processing is applied. In serial or parallel environments, the space complexity should be O$(N)$.

The time complexity of the program may be improved when it has been developed, as envisaged, so that the New patterns are processed in batches, with a purging of Old between each batch to remove all patterns except those in the best grammar. In this case, the time complexity should be O$(N)$.

\section{Example}\label{example_section}

When New contains the eight sentences shown in Figure \ref{example_2_patterns}, the best grammar found by SP70 is the one shown in Figure \ref{example_2_grammar}.

\begin{figure}[!hbt]
\begin{center}
\begin{BVerbatim}
t h a t b o y r u n s
t h a t g i r l r u n s
t h a t b o y w a l k s
t h a t g i r l w a l k s
s o m e b o y r u n s
s o m e g i r l r u n s
s o m e b o y w a l k s
s o m e g i r l w a l k s
\end{BVerbatim}
\end{center}
\caption{Eight sentences supplied to SP70 as New.}
\label{example_2_patterns}
\end{figure}

\begin{figure}[!hbt]
\begin{center}
\begin{BVerbatim}
< %2 2 s o m e >
< %2 3 t h a t >
< %1 5 b o y >
< %1 6 g i r l >
< %3 4 r u n s >
< %3 7 w a l k s >
< 1 < %2 > < %1 > < %3 > >
\end{BVerbatim}
\end{center}
\caption{The best grammar (in terms of MLE principles) that is found by SP70 when New contains the eight sentences shown in Figure \ref{example_2_patterns}.}
\label{example_2_grammar}
\end{figure}

\subsection{Intermediate Results}

As the first phase of learning proceeds (operation 3 of Figure \ref{SP70_figure}), intermediate results are often much less tidy than the example shown in Section \ref{derive_patterns_section}. For example, when Old contains only the first pattern shown in Figure \ref{example_2_patterns}, the only alignment it can create is:

\begin{center}
\begin{BVerbatim}
0 t h a  t b o y r u n s         0
         |                      
1 < %1 9 t h a t b o y r u n s > 1
\end{BVerbatim}
\end{center}

Notice that the Old pattern (in row 1) is, in effect, {\em the same pattern} as the New pattern (in row 0) so it is not permissible to match `o' in the New pattern, for example, with `o' in the Old pattern because that would mean matching a given symbol with itself!

From the alignment just shown, the program derives `bad' patterns like `$<$ \%3 14 t h a $>$', `$<$ \%4 18 b o y r u n s $>$' and `$<$ \%4 17 h a t b o y r u n s $>$' and these are added to Old. However, as later patterns are processed, the repository of Old patterns begins to accumulate enough patterns that are good in MLE terms so that it is able to create quite respectable looking parsings like this:

\begin{center}
\fontsize{07.00pt}{08.40pt}
\begin{BVerbatim}
0                    t h a t                g i r l                w a l k s     0
                     | | | |                | | | |                | | | | |
1                    | | | |                | | | |   < %8 %36 926 w a l k s >   1
                     | | | |                | | | |   | |                    |
2 < %10 220 < %7     | | | | > < %9         | | | | > < %8                   > > 2
            | |      | | | | | | |          | | | | |                       
3           | |      | | | | | < %9 %18 215 g i r l >                            3
            | |      | | | | |                                              
4           < %7 208 t h a t >                                                   4
\end{BVerbatim}
\end{center}

In the {\em sifting\_and\_sorting()} phase, all the `bad' patterns are discarded and the `good' patterns are cleaned up by removing unnecessary ID-symbols and renaming the retained ID-symbols in a tidy manner.

\subsection{Values for $G$, $E$, $T$ and Compression}

Figure \ref{plotting_table} shows changing values for $G$, $E$ and $T$ for the best grammar found (in terms of MLE principles) as successive patterns from New are processed in {\em compile\_alternative\_grammars()}. It is interesting to see that, as successive patterns are processed, progressively more compression is achieved, represented by the falling values for ($T$ / `original'), shown in the last column. 

\setlength{\tabcolsep}{2mm}
\begin{table}[!hbt]
\begin{center}
\begin{tabular}{| r | r | r | r | r | r |} \hline
\em Pattern & \em G &     \em E & \em T   &   \em Original &  \em Compression \\ \hline\hline
1 &           7970.49 &   26.78 &     7997.27 &   7943.70 &       1.00 \\
2 &           11085.38 &  191.29 &    11276.67 &  16569.42 &      0.68 \\
3 &           14665.26 &  302.09 &    14967.35 &  25195.14 &      0.59 \\
4 &           14665.26 &  397.57 &    15062.83 &  34502.87 &      0.44 \\
5 &           17650.07 &  563.32 &    18213.39 &  42488.08 &      0.42 \\
6 &           17650.07 &  713.75 &    18363.82 &  51155.30 &      0.36 \\
7 &           17650.07 &  887.00 &    18537.07 &  59822.52 &      0.31 \\
8 &           17650.07 &  1044.92 &   18694.99 &  69171.76 &      0.27 \\ \hline
\end{tabular}
\end{center}
\caption{Cumulative values (in bits) of $G$, $E$ and $T$ for the best grammar found as successive patterns from New are processed in {\em compile\_alternative\_grammars()}. For comparison purposes, the cumulative sizes of the original patterns (excluding ID-symbols) are shown in the `original' column and values for compression ($T$ / `original') are shown in the last column.}
\label{plotting_table}
\end{table}

\section{Discussion}\label{discussion_section}

\subsection{Evaluation}

In accordance with the `looks-good-to-me' approach to the evaluation of grammar induction systems \cite{van_zaanen_thesis_2002}, the grammar shown in Figure \ref{example_2_grammar} looks like an appropriate grammar for the patterns shown in Figure \ref{example_2_patterns}.\footnote{A possible improvement might be a grammar that isolates the `s' in `r u n s' and `w a l k s' as a separate morpheme.} This may seem like a sloppy method of evaluation but it should not be forgotten that the human brain is, by a wide margin, the best learning system on the planet. This provides a justification for using human judgement of what does or does not `look good' as a means of evaluating the output of artificial learning systems. With any system that is sufficiently robust to be applied to realistic samples of natural language, then there is no alternative to (human) judgements about what is or is not a `correct' grammar for a given language or (human) conventions about how language is segmented into words. Statistical tests may be applied to establish whether or not there is a significant level of agreement between structures established by human judgement and the results of artificial learning \cite{wolff_1977,wolff_1980}.

Notice that the use of a `target' grammar as a criterion of success (as in Gold's approach to learning \cite{gold_1967}) does not overcome the problem that, for any given language sample, there are many alternative grammars that are compatible with the sample and some are `better' than others.

\subsection{Reorganisation Needed}

The example in the previous section is good enough to show that the approach is sound but experiments with other examples have shown that the model suffers from two main weaknesses:

\begin{itemize}

\item Although the model in its current form can isolate basic segments and tie them together in an overall abstract structure, it is not good at finding intermediate levels of abstraction.

\item In the development of the model to date, no attempt has been made to enable the system to detect discontinuous dependencies such as number dependency between the subject of a sentence and its main verb (as mentioned in Section \ref{multiple_alignment_section}). Although this kind of capability may seem like a refinement that we can afford to do without at this stage of development, a deficiency in this area seems to have an impact on the program's performance at an elementary level.

\end{itemize}

A possible solution to both problems is a reorganisation of the model so that learning is integrated even more closely with parsing. Recent work has shown that operation 2.2 in the {\em sifting\_and\_sorting()} function (Figure \ref{sifting_and_sorting_figure}) can be omitted---the multiple alignments from operation 3.2 in Figure \ref{SP70_figure} can be used instead. It is also envisaged that New patterns will be processed in batches and that, after each batch, {\em sifting\_and\_sorting()} will be applied and Old patterns that are not proving useful will be discarded.

\section{Conclusion}

SP70 is not yet an `industrial strength' system for unsupervised learning but I believe the framework has considerable potential and provides a sound basis for further development.

A key attraction of this approach to learning is that the ICMAUS framework provides a unified view of a variety of issues in AI thus facilitating the integration of grammar induction with other aspects of intelligence. Given the generality of the framework, the learning techniques described here are relevant to the learning of {\em any} kind of knowledge, not just grammars.

\section*{Acknowledgements}

I am grateful to Pat Langley and Menno van Zaanen for constructive comments on the report on which this paper is based. The paper has also benefitted from comments and suggestions by an anonymous referee.

\raggedright

\end{document}